\documentclass[runningheads]{llncs}
\usepackage[T1]{fontenc}
\usepackage{multicol}
\usepackage{multirow}
\usepackage{graphicx,verbatim}
\usepackage{amssymb}
\usepackage{booktabs}
\usepackage{comment}
\usepackage{float}
\usepackage{makecell}
\usepackage{amsmath}
\usepackage{adjustbox}
\usepackage{color}

\usepackage{hyperref}
\hypersetup{
    colorlinks=true,
    linkcolor=blue,
    citecolor=blue,
    urlcolor=blue,
    }

\begin{document}

\title{Disentangled Shared Representations Improve Morpho-Transcriptomic Integration}
\titlerunning{Disentangled shared representations in ST}
\begin{comment}  
 Removed for anonymized MICCAI submission
\author{First Author\inst{1}\orcidID{0000-1111-2222-3333} \and
Second Author\inst{2,3}\orcidID{1111-2222-3333-4444} \and
Third Author\inst{3}\orcidID{2222--3333-4444-5555}}

\authorrunning{F. Author et al.}
\institute{Princeton University, Princeton NJ 08544, USA \and
Springer Heidelberg, Tiergartenstr. 17, 69121 Heidelberg, Germany
\email{lncs@springer.com}\\
\url{http://www.springer.com/gp/computer-science/lncs} \and
ABC Institute, Rupert-Karls-University Heidelberg, Heidelberg, Germany\\
\email{\{abc,lncs\}@uni-heidelberg.de}}

\end{comment}

\author{
Julian Ostermaier\inst{1,2}$^{*}$
\and
Swann Ruyter\inst{2}
\and
Reuben Dorent\inst{2}
\and
Daniel Racoceanu\inst{2}
}

\authorrunning{J. Ostermaier et al.}

\institute{
ESPCI Paris, PSL University, Paris, France\\
\email{julian.ostermaier@gmx.net}
\and
Sorbonne Universit\'e, CNRS, Inserm, AP-HP, Inria, Paris Brain Institute -- ICM, Paris, France\\
\email{swann.ruyter@icm-institute.org},
\email{reuben.dorent@inria.fr},
\email{daniel.racoceanu@sorbonne-universite.fr}\\
$^{*}$Corresponding author. \quad
}
  
\maketitle              % typeset the header of the contribution
\begin{abstract}
Spatial transcriptomics (ST) enables the simultaneous profiling of gene expression and tissue morphology, creating an opportunity to learn multimodal representations capturing shared morpho-transcriptomic structure. However, standard multimodal models often compress modalities into a common latent space without explicitly separating shared and modality-specific sources of variation, which may limit downstream utility. We investigate whether explicit disentanglement of shared and private latent components improves multimodal representation learning for paired Hematoxylin \& Eosin (H\&E) and ST data. We compare VAE-based and contrastive  approaches, each in standard and disentangled variants, across two cancer cohorts under matched experimental conditions. Representations are evaluated using cross-modal reconstruction, downstream probing and cross-modal probe transfer.  The experiments suggest two main trends. First, contrastive objectives yield higher downstream probing performance than VAE-based models. Second, disentangled variants improve the selected reconstruction and probing metrics, although the gains depend on the model family, task, direction, and disentanglement strength. Overall, our results suggest that explicitly factorizing shared and modality-specific information can improve multimodal representation learning for spatial transcriptomics and provides a useful evaluation framework for future foundation models.

% Authors must provide keywords and are not allowed to remove this Keyword section.

\end{abstract}
\section{Introduction}

Spatial transcriptomics (ST) enables spatially resolved molecular profiling while preserving tissue architecture, providing paired measurements of histology and gene expression from the same tissue section \cite{2Moses2022-ye_museum}.  Beyond direct prediction of gene expression from H\&E ~\cite{2zhu2025diffusion,2Tran2026-bd}, these paired data offer an opportunity to learn multimodal representations of tissue organization.

In histopathology, foundation models have demonstrated that compact image representations capture transferable tissue features across diverse tasks and cohorts \cite{2lu2024avisionlanguage,2chen2024uni}. A similar objective is appealing for ST, where joint representations of H\&E morphology and gene expression can encode the morpho-transcriptomic organization of a tissue section in a low-dimensional latent space. Recent advances in representation learning, including in world models, suggest that learning compact latent representations of complex data can provide a more efficient alternative to directly modeling high-dimensional observations \cite{maes2026leworldmodelstableendtoendjointembedding}. In the context of ST, this motivates learning latent representations that capture morpho-transcriptomic information, rather than directly predicting virtual gene expression  from H\&E. Such representations may provide compact transcriptomics-enriched representations for downstream analyses and transfer tasks.

Accordingly, several multimodal representation learning methods have recently been proposed to learn a shared representation of these modalities using paired histology and ST data. Approaches range from contrastive objectives~\cite{2xie2023spatially,2Chen2025} to autoencoder-based and generative models~\cite{2DAIAsthetik,2Bao2022Muse,2wood2025genst,2spatialDIVA}. However, while standard multimodal models encode both modalities into the same space, they do not control which information is \emph{shared} and which is \emph{private} (or modality-specific). This disentanglement is needed in representation learning \cite{2spatialDIVA,2Chelebian2025}, since the true amount of shared signal between H\&E and gene expression data is unknown and both modalities contain a large amount of modality-specific information. The idea of disentangling shared from modality-specific information is well established in the broader multimodal learning literature \cite{2palumbo2023mmvae,2pmlr-v240-martens24a,2wang2025an}. Within ST specifically, ~\cite{2spatialDIVA} is the closest prior work: it uses a latent variable model that explicitly decomposes histological and ST data into distinct generative factors including spatial, morphological, and transcriptomic variation. However, it relies on expert annotations for supervision which makes it impractical for large-scale training, where such annotations are costly and often unavailable. More broadly, across this line of work, previously proposed representation learning approaches are evaluated under heterogeneous experimental settings and different evaluation criteria, leaving it unclear if their representations are both well aligned and simultaneously keep information of downstream utility.

In this work, we test whether disentanglement of multimodal representations improves joint H\&E and ST representation learning. We compare two  main lines of models, variational autoencoder models (VAE) and contrastive  models, each in standard shared-latent and variants that separate shared from modality-specific components \cite{2palumbo2023mmvae,2pmlr-v240-martens24a,2wang2025an}. Under matched experimental conditions, we evaluate the learned representations across multiple tissue cohorts using cross-modal reconstruction, downstream probing, and cross-modal probe transfer. Our experiments suggest that contrastive objectives yield stronger downstream representations than VAE-based models, while disentanglement generally improves cross-modal prediction and downstream performance.

\section{Methods}
In this study, we compared five representative multimodal representation learning models, including multimodal variational autoencoders and contrastive learning methods, and their disentangled extensions, across two cancer cohorts, using a unified probing and reconstruction evaluation framework. 

\subsection{Representation Learning Model Baselines.}
Representation methods were selected to represent two common families of multimodal representation learning methods: autoencoder-based objectives that learn shared representations through cross-modal reconstruction, as in GenST \cite{2wood2025genst}, and contrastive alignment objectives, as exemplified by omiCLIP \cite{2Chen2025}. Crucially, to explore the effect of disentanglement, both lines were represented by a standard shared-representation model and a disentangled variant, enabling a systematic comparison of representation strategies within each paradigm (see Table \ref{tab:model_summary}).  \\

    \noindent \textbf{Multimodal Variational Autoencoders (MMVAE) \cite{2MMVAE_SHI}.}  MMVAE encodes each modality independently into a shared latent space. 
    The mixture-of-experts VAE~\cite{2MMVAE_SHI}, which approximates the joint posterior distribution using a mixture-of-experts, serves as the non-disentangled MMVAE baseline.  We further implemented \textbf{MMVAE}+\cite{2palumbo2023mmvae}, a MMVAE extension that introduces an explicit disentanglement mechanism between shared and modality-specific information.
    % The encoders inferring the shared representations $\mathbf{z}_s$ are optimized (\ref{mmvaeloss}) for cross-modal reconstruction, while the modality-private latents $\mathbf{z}_p$ are only used for self-reconstruction.
    We denote the modalities by $\mathcal{M}=\{G,\mathrm{HE}\}$, where $\mathbf{x}_m$ is the observation from modality $m$. The shared latent code is $\mathbf{z}_s$, and $\mathbf{z}_p^m$ denotes the private latent code of modality 
    $m$. We write $\mathbf{z}_p=\{\mathbf{z}_p^m\}_{m\in\mathcal{M}}$. 
    The encoders $q_s^m$ and $q_p^m$ approximate the modality-specific posteriors over shared and private latents, respectively. Its objective can be written as
    \begin{equation}
    \mathcal{L}_{\mathrm{MMVAE+}}
    =
    \frac{1}{|\mathcal{M}|}
    \sum_{m \in \mathcal{M}}
    \mathbb{E}_{\substack{q_s^m(\mathbf{z}_s \mid \mathbf{x}_m) \\ q_p^m(\mathbf{z}_p^m \mid \mathbf{x}_m)}}
    \left[
    \log \left(
    \frac{
    R_{mm}
    }{
    q_\Phi(\mathbf{z}_s,\mathbf{z}_p \mid \mathbf{x}_{1:M})
    }
    \prod_{\substack{n \in \mathcal{M}, \\ n \neq m}}
    R_{mn}
    \right)
    \right],
    \label{mmvaeloss}
    \end{equation}
    where
    $R_{mm} = p_m(\mathbf{x}_m,\mathbf{z}_s,\mathbf{z}_p^m)$ is self-reconstruction,
    $R_{mn} = p_{\theta_n}(\mathbf{x}_n \mid \mathbf{z}_s,\tilde{\mathbf{z}}_p^n)$ cross-modal reconstruction and $\tilde{\mathbf{z}}_p^n \sim r(\mathbf{z}_p^n)$ is sampled from a modality-specific prior distribution. Additionally, the recent disentanglement variant \textbf{MMVAE+$\mathrm{sg}$} \cite{2pmlr-v240-martens24a} was included. \textbf{MMVAE+$\mathrm{sg}$} redefines the training objective to prevent leakage of modality-specific information into the shared space, using a stop-gradient operator that blocks the gradient flow from same-view reconstruction into the shared latent variable $\mathbf{z}_s$. \\

    \noindent \textbf{Contrastive Learning (CLIP)~\cite{2radford2021learningtransferablevisualmodels}. } 

    Motivated by recent  foundation models for ST such as 
omiCLIP~\cite{2Chen2025} and BLEEP~\cite{2xie2023spatially}, 
contrastive learning methods were included as a second model family. Modality-specific 
encoders were trained with the symmetric InfoNCE 
loss~\cite{2radford2021learningtransferablevisualmodels}, which maximizes 
the cosine similarity between representations of paired multimodal observations while minimizing the similarity to independent observations 
within each training batch, thereby aligning 
representations across modalities without requiring data reconstruction:
\begin{equation}
\mathcal{L}_{\mathrm{con}} = 
-\frac{1}{2}\mathbb{E}\left[
\log \frac{\exp(\mathbf{z}_s^{G\top} \mathbf{z}_s^{\mathrm{HE}+} / \tau)}
{\sum_{k} \exp(\mathbf{z}_s^{G\top} \mathbf{z}_{s,k}^{\mathrm{HE}} / \tau)}
+ \log \frac{\exp(\mathbf{z}_s^{\mathrm{HE}\top} \mathbf{z}_s^{G+} / \tau)}
{\sum_{k} \exp(\mathbf{z}_s^{\mathrm{HE}\top} \mathbf{z}_{s,k}^{G} / \tau)}
\right],
\label{eq:infonce}
\end{equation}
where $\tau$ is a temperature parameter, superscript $+$ denotes the 
matched (positive) sample, and the sums run over all samples in the batch.

\noindent \textbf{Disentangled SSL (disSSL)}~\cite{2wang2025an} extends this contrastive framework by separating shared from modality-specific latent information. The shared representation is learned using the same contrastive objective $\mathcal{L}_{\mathrm{con}}$ with a penalty that minimizes $I\left( \mathbf{Z}_s^{G}, \mathbf{X}_{\mathrm{HE}}\mid \mathbf{X}_{G}\right)$, the amount of modality-specific information retained in $\mathbf{z}_s$, using the Kullback-Leibler divergence ($D_{KL}$) as an upper bound:    
    \begin{equation}
    \mathcal{L}_{\mathbf{z}_s}
    =
    \underbrace{\mathcal{L}_{\mathrm{con}}}_{\text{contrastive loss}}
    +
    \beta\,
    % \underbrace{D_{KL}(q_{_s^{G}}(\mathbf{z}_{_s^{G}} \mid \mathbf{x}_G) \mid \mid q_{_s^{HE}}(\mathbf{z}_{_s^{HE}} \mid \mathbf{x}_{HE}))}_{\text{redundancy reduction}} 
    \underbrace{D_{KL}
    (q_s^G(\mathbf{Z}_s^{G}\mid \mathbf{X}_G) \mid\mid q_s^{\mathrm{HE}}(\mathbf{Z}_s^{\mathrm{HE}}\mid \mathbf{X}_{\mathrm{HE}}))}_{\text{redundancy reduction}}.
    \label{eq:dissl}
    \end{equation}
    % In this formulation, $\mathcal{L}_{\mathrm{con}}$ denotes the contrastive alignment objective, 
    %while $\mu(\mathbf{x}_G)$ and $\mu_{\mathrm{HE}}(\mathbf{x}_{\mathrm{HE}})$ denote modality-specific mean representations entering the redundancy penalty. 
    % while the second term is the  Kullback-Leibler divergence between the distributions of the shared representations inferred from each modality.
    The coefficient $\beta$ controls the strength of this penalty. When $\beta=0$, the objective reduces to standard contrastive alignment. \\

    \noindent \textbf{Uni-modal Baselines.} To contextualize the performance of our multimodal models, uni-modal baselines were included using Principal Component Analysis (PCA) and uni-modal Variational Autoencoders (VAE), thereby providing linear and nonlinear reference representation baselines within each modality.\\

\noindent \textbf{Implementation Details.} All models were trained using a leave-one-slide-out cross-validation scheme to learn a 32-dimensional shared latent representation~$\mathbf{z}_s$, capturing information shared across gene expression and histology modalities.
Architectural choices and training settings were matched across models where applicable.
Experiments were implemented in Python with \texttt{PyTorch} and run on NVIDIA V100 GPUs. MMVAE-based models were adapted from \texttt{MultiVae} ~\cite{2Senellart2025}, with preprocessing performed using \texttt{HEST} and \texttt{scanpy}. Probing experiments were repeated five times with different random seeds; training hyperparameters are reported in Supplementary Table~\ref{tab:hyperparams}.

\begin{figure}
\includegraphics[width=\textwidth]{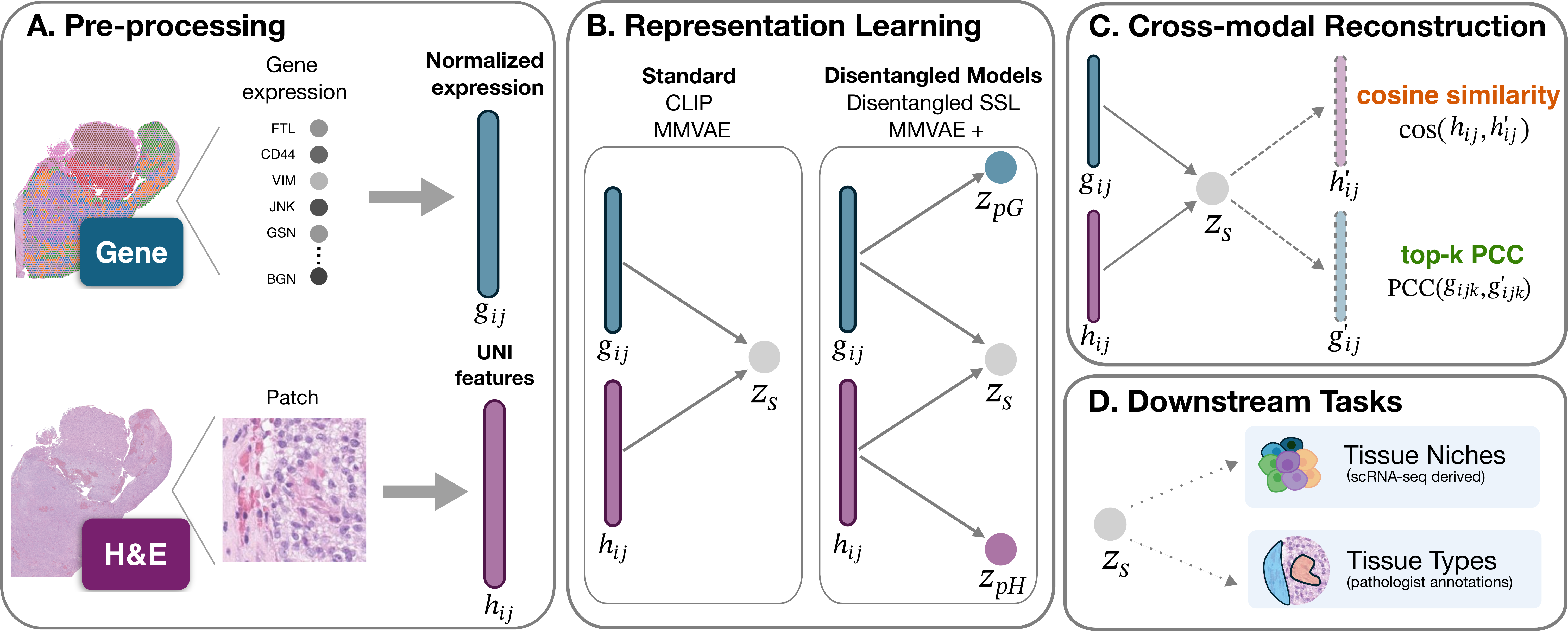}
\caption{\textbf{Multimodal representation learning pipeline.} \textbf{(A)} Pre-processing of the paired gene expression and H\&E image patch data. \textbf{(B)} Representations are learned using different model families including contrastive and VAE methods as well as their disentangled versions. Representations are evaluated using \textbf{(C)} cross-modal reconstruction and \textbf{(D)} prediction of downstream labels.} \label{pipeline}
\end{figure}

\subsection{Datasets and Pre-processing}
\noindent \textbf{Cohorts.}
We used two Visium ST cohorts: 7 colorectal cancer (Colon) samples from HEST~\cite{2jaume2024hest} and 10 glioblastoma (GBM) samples acquired in this project from \textit{Owkin - MOSAIC Window initiative}. For Colon, one sample per patient was retained, selected based on artifact burden and spot count, to avoid patient overlap during leave-one-slide-out cross-validation. The GBM cohort complements Colon by introducing stronger inter-patient molecular heterogeneity, a known feature of GBM~\cite{2Wick2018}.  After pre-processing this results in a total of 35,002 spots for the GBM and 62,730 spots for the Colon cohort.\\

\noindent \textbf{Spatial Transcriptomics Pre-processing.}
Gene expression data were filtered by retaining spots with more than 200 expressed genes and fewer than 8\% mitochondrial counts, normalized to 10,000 counts per spot, and log-transformed. For each dataset, gene panels were obtained from the union of highly variable genes (HVGs) across samples. HVGs were ranked by their maximum variance across slides, and the top $K$ genes were retained ($K \in \{200, 500, 1000\}$). \\

\noindent \textbf{Histology Image Pre-processing.}
Histology image patches were extracted centered on each Visium spot, with a patch size of 224 × 224 pixels at a resolution of 0.5 microns per pixel, fully encompassing the 55 µm Visium spot diameter. Patches were processed with the UNI histopathology foundation model~\cite{2chen2024uni} to obtain a 1024-dimensional image embedding for each spot, capturing morphologically relevant features from the image modality.

\subsection{Experimental Setup}
Each model was trained to map multimodal inputs into a low-dimensional joint representation space. Once trained, the encoder weights were frozen and the learned representations were evaluated using probing models with a fixed  architecture (Figure~\ref{pipeline}) to isolate the quality of the learned representations. Evaluation comprised cross-modal reconstruction and downstream prediction tasks. Probing on each task was repeated five times with different random seeds. \\

\noindent \textbf{Cross-modal Reconstruction Task.}
Cross-modal reconstruction evalu\-ates if a shared representation obtained from one modality can be used to predict the other modality. 

Specifically, MLP decoders were trained to reconstruct gene expression from the H\&E-conditioned representation $\mathbf{z}_s^{\mathrm{HE}}$ (HE$\rightarrow$G), and image representations from the gene-expression-conditioned representation $\mathbf{z}_s^{G}$ (G$\rightarrow$HE). HE$\rightarrow$G reconstruction is evaluated using top-50 PCC, whereas G$\rightarrow$HE reconstruction is evaluated using cosine similarity. Top-50 PCC is defined as the mean PCC across the 50 best-predicted genes, while cosine similarity is computed between the predicted and ground-truth image embeddings.  Results are reported in Table~\ref{tab:recon_all}. \\

\noindent \textbf{Downstream Probing Task.}

We evaluated downstream utility on the Colon cohort using linear probes for two 
complementary label sets from~\cite{2Valdeolivas2024-rr}: pathologist tissue-type annotations and tissue niches derived from scRNA-seq-based cell-type composition estimates. Tissue-type labels provide a morphology-oriented task, expected to be most directly captured by H\&E-conditioned representations, whereas tissue-niche labels provide a molecularly oriented task, expected to be most directly captured by gene-expression-conditioned representations. Label construction is described in Supplementary Section~\ref{sup:downstream}.

\begin{figure}[t]
\includegraphics[width=\textwidth]{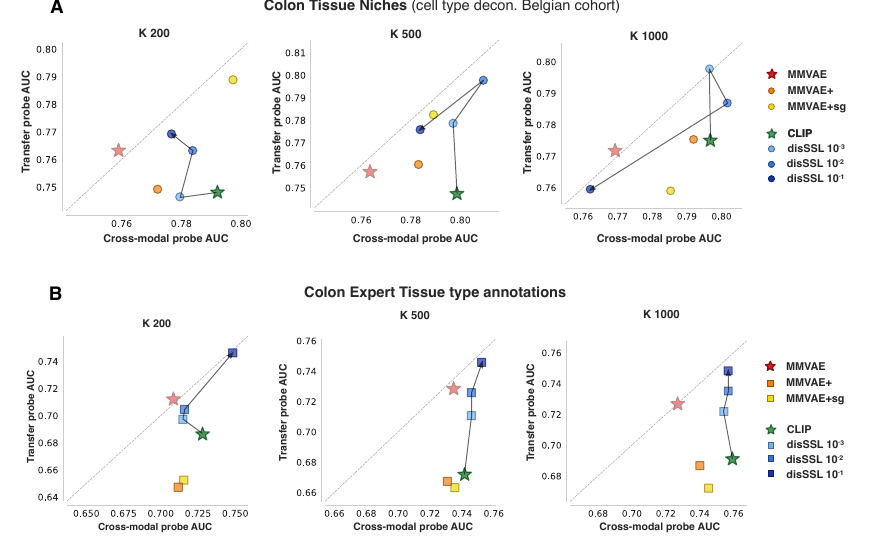}
\caption{\textbf{Downstream task performance for tissue niche classification ($HE \rightarrow G$) and tissue type classification ($G \rightarrow HE$).}  AUC prediction performance (five repeated runs with random seeds) with gene panel sizes $K$ for \textbf{(A)} tissue niche classification derived from scRNA-seq reference Belgian cohort and \textbf{(B)} pathologist tissue types classification. \textit{Stars} indicate non-disentangled and \textit{squares} disentangled versions. The gray arrow connects contrastive models from CLIP ($\beta=0$) to increasing disentanglement strength.} \label{downstream_colon_niches}
\end{figure}

We report two probing settings. 1) Cross-modality probes evaluate whether representations conditioned on one modality predict labels primarily associated with the other modality: HE$\rightarrow$G probes predict tissue niches from H\&E-conditioned representations, whereas G$\rightarrow$HE probes predict tissue-type annotations from gene-expression-conditioned representations.  2) Cross-modality transfer probing assesses if a probe trained on one modality-conditioned representation can be applied to the other without retraining. This allows to assess if the shared latent space only contains shared information and if this information is rich. AUC results are reported for tissue niche classification and tissue-type classification in Figure~\ref{downstream_colon_niches}.

\section{Results}
This section presents the cross-modal and downstream results of our study. \\

\noindent \textbf{Cross-modal Reconstruction Results.} Within the VAE family, MMVAE+ and MMVAE$+_{\mathrm{sg}}$ achieved higher mean  reconstruction performance than MMVAE across both cohorts, panel sizes, and reconstruction directions. After multiple-test correction, only a subset of these comparisons reached statistical significance (Table~\ref{tab:stat}), indicating that this pattern should be interpreted as a descriptive trend rather than a uniform effect.

Among contrastive models, disSSL showed a direction- and $\beta$-dependent pattern. For HE$\rightarrow$G reconstruction, the best disSSL configuration exceeded CLIP on GBM at larger panel sizes ($K=500$ and $K=1000$), but not at $K=200$. On Colon, disSSL matched or slightly exceeded CLIP at $K=200$ and $K=1000$, whereas CLIP remained higher at $K=500$. For G$\rightarrow$HE reconstruction, disSSL did not consistently improve over CLIP. This asymmetry is consistent with stronger disentanglement removing information useful for image-embedding reconstruction, rather than uniformly improving cross-modal prediction. Because most comparisons did not reach significance after correction, these patterns should be interpreted as suggestive. \\

% Reconstruction table macros
\newcommand{\crval}[2]{{#1$\pm$#2}}
\newcommand{\crbest}[2]{\textbf{#1$\pm$#2}}
\newcommand{\crsecond}[2]{\underline{#1$\pm$#2}}

\begin{table*}[t]

\centering
\scriptsize
\setlength{\tabcolsep}{2.7pt}
\renewcommand{\arraystretch}{1.10}
\caption{
Cross-modal reconstruction results on Colon and GBM datasets for different gene panel sizes $K$.
Metrics are reported as percentages and presented as mean$\pm$standard deviation over five repeated runs with different random seeds.
For disSSL, $10^{-3}$, $10^{-2}$, and $10^{-1}$ denote the value of $\beta$.
\textbf{Bold} and \underline{underline} indicate the best and second-best values per column within each model family.
}
\label{tab:recon_all}

\begin{tabular}{@{}lllcccccc@{}}
\toprule
\textbf{} &
\textbf{Family} &
\textbf{Model} &
\multicolumn{3}{c}{\textbf{top-50 PCC} (HE$\to$G)} &
\multicolumn{3}{c}{\textbf{cosine similarity} (G$\to$HE)} \\
\cmidrule(lr){4-6}
\cmidrule(l){7-9}
& & &$K$ 200 & 500 & 1000 &$K$ 200 & 500 & 1000 \\
\midrule

\multirow{9}{*}{\textbf{Colon}}
& \multirow{2}{*}{\begin{tabular}[c]{@{}l@{}}Uni-\\modal\\\end{tabular}}
& PCA
& \crval{47.9}{1.5}
& \crval{51.4}{1.0}
& \crval{52.7}{0.8}
& \crval{59.2}{1.0}
& \crval{60.7}{0.5}
& \crval{60.7}{0.8} \\

& 
& VAE
& \crval{43.7}{0.6}
& \crval{46.7}{0.8}
& \crval{47.2}{1.0}
& \crval{57.0}{0.5}
& \crval{58.5}{0.4}
& \crval{58.9}{0.5} \\

\cmidrule(lr){2-9}

& \multirow{3}{*}{VAE}
& MMVAE
& \crval{42.3}{0.4}
& \crval{48.6}{0.2}
& \crval{49.7}{0.5}
& \crval{60.0}{0.3}
& \crval{59.9}{0.3}
& \crval{61.4}{0.2} \\

&
& MMVAE+
& \crsecond{47.5}{0.6}
& \crsecond{50.8}{0.8}
& \crsecond{52.5}{0.8}
& \crsecond{61.5}{0.4}
& \crsecond{60.8}{0.3}
& \crsecond{62.5}{0.5} \\

&
& MMVAE+$\mathrm{sg}$
& \crbest{47.6}{0.5}
& \crbest{51.8}{0.4}
& \crbest{53.2}{0.6}
& \crbest{62.3}{0.3}
& \crbest{61.5}{0.5}
& \crbest{62.7}{0.4} \\

\cmidrule(lr){2-9}

& \multirow{4}{*}{\begin{tabular}[c]{@{}l@{}}Contras-\\tive\end{tabular}}
& CLIP
& \crsecond{50.2}{0.5}
& \crbest{53.4}{0.5}
& \crval{52.8}{0.6}
& \crbest{63.1}{0.3}
& \crbest{62.0}{0.3}
& \crsecond{62.4}{0.2} \\

&
& disSSL $10^{-3}$
& \crbest{50.6}{0.8}
& \crsecond{53.1}{0.3}
& \crsecond{53.3}{0.5}
& \crsecond{61.8}{0.5}
& \crsecond{61.9}{0.5}
& \crbest{62.5}{0.5} \\

&
& disSSL $10^{-2}$
& \crval{48.8}{0.4}
& \crval{52.3}{0.9}
& \crbest{53.9}{0.4}
& \crval{61.3}{0.7}
& \crval{60.8}{0.9}
& \crval{62.2}{0.5} \\

&
& disSSL $10^{-1}$
& \crval{45.7}{0.3}
& \crval{46.5}{0.3}
& \crval{47.4}{0.6}
& \crval{60.9}{0.9}
& \crval{60.9}{0.9}
& \crval{60.8}{0.4} \\

\midrule

\multirow{9}{*}{\textbf{GBM}}
& \multirow{2}{*}{\begin{tabular}[c]{@{}l@{}}Uni-\\modal\\\end{tabular}}
& PCA
& \crval{43.0}{0.3}
& \crval{46.9}{0.8}
& \crval{47.4}{0.7}
& \crval{38.7}{0.5}
& \crval{40.7}{0.4}
& \crval{41.6}{0.3} \\

&
& VAE
& \crval{42.2}{0.4}
& \crval{43.6}{0.7}
& \crval{43.5}{0.7}
& \crval{39.8}{0.1}
& \crval{39.5}{0.4}
& \crval{40.0}{0.3} \\

\cmidrule(lr){2-9}

& \multirow{3}{*}{VAE}
& MMVAE
& \crval{46.1}{0.4}
& \crval{50.0}{0.3}
& \crval{51.7}{0.2}
& \crval{41.9}{0.3}
& \crval{42.6}{0.3}
& \crsecond{44.6}{0.2} \\

&
& MMVAE+
& \crsecond{48.3}{0.3}
& \crsecond{51.0}{0.2}
& \crbest{52.3}{0.5}
& \crbest{43.7}{0.2}
& \crbest{44.4}{0.3}
& \crbest{45.0}{0.1} \\

&
& MMVAE+$\mathrm{sg}$
& \crbest{48.0}{0.3}
& \crbest{51.4}{0.3}
& \crsecond{51.8}{0.3}
& \crsecond{42.8}{0.4}
& \crsecond{43.9}{0.2}
& \crval{44.5}{0.5} \\

\cmidrule(lr){2-9}

& \multirow{4}{*}{\begin{tabular}[c]{@{}l@{}}Contras-\\tive\end{tabular}}
& CLIP
& \crbest{48.6}{0.2}
& \crval{50.1}{0.2}
& \crval{50.8}{0.2}
& \crbest{42.9}{0.4}
& \crsecond{44.4}{0.6}
& \crbest{45.7}{0.7} \\

&
& disSSL $10^{-3}$
& \crsecond{48.3}{0.2}
& \crsecond{50.5}{0.5}
& \crsecond{51.0}{0.3}
& \crsecond{42.5}{0.3}
& \crbest{45.0}{0.4}
& \crsecond{44.3}{0.2} \\

&
& disSSL $10^{-2}$
& \crval{47.9}{0.3}
& \crbest{51.0}{0.5}
& \crbest{51.7}{0.3}
& \crval{41.6}{0.5}
& \crval{44.0}{0.7}
& \crval{43.8}{0.5} \\

&
& disSSL $10^{-1}$
& \crval{41.8}{0.2}
& \crval{44.8}{0.5}
& \crval{46.0}{0.3}
& \crval{38.5}{0.6}
& \crval{38.4}{0.8}
& \crval{40.0}{0.6} \\

\bottomrule
\end{tabular}
\end{table*}

\noindent \textbf{Downstream Probing Results.} 

Contrastive models achieved higher cross-modal probe AUCs than VAE-based models for both tissue niche and tissue-type classification. MMVAE showed the lowest cross-modality probe performance, whereas MMVAE+ and MMVAE$+_{\mathrm{sg}}$ improved over MMVAE but generally remained below CLIP and disSSL. This pattern suggests that, under this probing protocol, contrastive alignment yields more task-useful shared representations than VAE-based reconstruction objectives.

Within the VAE family, the benefit of disentanglement depended on the probing setting. MMVAE+ and MMVAE$+_{\mathrm{sg}}$ improved cross-modality probing across both label sets, supporting a benefit of factorized latent representations for this readout. Transfer probing showed a more task-dependent pattern: MMVAE$+_{\mathrm{sg}}$ improved transfer performance for tissue niche classification in several settings, whereas disentangled VAE variants showed lower transfer AUCs for tissue-type annotations. This reversal suggests that the shared latent representation may retain less morphology-associated information when the VAE objective is more strongly factorized.

Among contrastive models, disSSL matched or improved over CLIP in several probing settings, but the effect depended on $\beta$, panel size, and task. For cross-modality probes, disSSL generally matched or exceeded CLIP up to an intermediate disentanglement strength, with weaker gains at the smallest panel size ($K=200$). For transfer probes, disSSL improved over CLIP in several conditions, consistent with improved latent-space compatibility and/or better retention of task-relevant information. However, performance was sensitive to the choice of $\beta$: stronger disentanglement did not uniformly improve AUC, and the optimal value varied across tasks and panel sizes. These results suggest that disentanglement can improve selected downstream readouts, but its benefit is not uniform across objectives, tasks, and hyperparameter settings.

\section{Conclusion}

We evaluated VAE-based and contrastive multimodal representation learning models for paired H\&E and ST data, comparing standard shared latent models with disentangled variants under a common evaluation protocol. The results support two main observations. First, disentangled VAE variants improved mean  cross-modal reconstruction performance relative to MMVAE, although statistically significant gains were limited to a subset of configurations. Second, contrastive models achieved higher downstream linear-probe AUCs than VAE-based models on the Colon classification tasks. In contrast, the benefit of disSSL over CLIP was more variable and depended on the task, panel size, and disentanglement strength. Together, these findings suggest that disentanglement can improve selected readouts of multimodal representation quality, but does not provide a uniform benefit across objectives or evaluation settings. They also highlight the need to combine cross-modal reconstruction, cross-modality probing, and transfer probing, as each criterion captures a different aspect of learned H\&E--ST representations.

\section*{Acknowledgments and Disclosure of Funding}
The research leading to these results has received funding from Agence Nationale de la Recherche as part of the “France 2030” program (reference ANR-23-IACL-0008, PRAIRIE-PSAI) and as part of the ”Investissements d’avenir” program (reference ANR-19-P3IA-0001, PRAIRIE 3IA Institute; and reference ANR-10-IAIHU-0006). The ARAMIS Lab is affiliated with DIM C-BRAINS, funded by the Conseil Régional d’Ile-de-France. This work was performed using HPC resources from GENCI–IDRIS (Grant 2025-AD011016416). R.D. received a Marie Sklodowska-Curie grant No 101154248 (project: SafeREG).

This study makes use of data generated by the MOSAIC consortium (Owkin, Charité – Universitätsmedizin Berlin, Lausanne University Hospital - CHUV, Erlangen Hospital, Gustave Roussy Institute, University of Pittsburgh) and made available through the MOSAIC Window initiative. Readers should note that the MOSAIC consortium bears no responsibility for the further analysis or interpretation of these data beyond what published by the MOSAIC consortium partners.

\subsubsection{Disclosure of Interests}
The authors have no competing interests in the paper.

\bibliographystyle{splncs04}
\bibliography{references}

\newpage

\section*{Supplementary Material}
\subsection*{Model Overview}
\begin{table}[h!]
\centering
\caption{Summary of the evaluated representation learning models.}
\label{tab:model_summary}
\small
\begin{tabular}{lcccc}
\toprule
\textbf{Model} & \textbf{Contrastive    } & \textbf{Generative    } & \textbf{Shared latent   } & \textbf{Private latent} \\
\midrule
MMVAE~\cite{2MMVAE_SHI}                  & -- & \checkmark & \checkmark & -- \\
MMVAE+~\cite{2palumbo2023mmvae}                  & -- & \checkmark & \checkmark & \checkmark \\
MMVAE+$\mathrm{sg}$~\cite{2pmlr-v240-martens24a}    & -- & \checkmark & \checkmark & \checkmark \\
CLIP~\cite{2radford2021learningtransferablevisualmodels}                    & \checkmark & -- & \checkmark & -- \\
disSSL~\cite{2wang2025an}                  & \checkmark & -- & \checkmark & \checkmark \\
\bottomrule
\end{tabular}
\end{table}
\subsection*{Statistical Tables}
\begin{table*}[h!]
\centering                         
\caption{Statistically significant Wilcoxon signed-rank test results after Benjamini--Hochberg FDR
correction ($p < 0.05$) across Colon and GBM dataset (HVG, log1p preprocessing). For each slide performance metrics were averaged across repeated runs, and paired slide-level metric values were used as observations in the Wilcoxon signed-rank test.
Significance levels refer to the BH-adjusted $p$-value:
$*\ p<0.05$;\; $**\ p<0.01$;\; $***\ p<0.001$.}
\label{tab:stat}
%
% ── Two-column layout: ccRCC+Colon on the left, GBM on the right 
\begin{tabular}{llccc}
\toprule
\textbf{Comparison} & \textbf{Metric} & \textbf{$p$ (BH)} & \textbf{Sig.} \\
\midrule
\multicolumn{4}{l}{\textbf{Colon}} \\
\midrule
---   & ---  & ---  \\
\bottomrule
\multicolumn{4}{l}{\textbf{GBM}} \\
\midrule
MMVAE vs MMVAE+sg           & PCC (HE$\to$G) 500     & 0.035 & $*$ \\
MMVAE vs MMVAE+             & Cosine (G$\to$HE) 200    & 0.015 & $*$ \\
MMVAE vs MMVAE+             & Cosine (G$\to$HE) 500    & 0.019 & $*$ \\

CLIP vs disSSL $\beta^{-3}$ & Cosine (G$\to$HE) 1000      & 0.019 & $*$   \\

CLIP vs disSSL $\beta^{-2}$ & Cosine (G$\to$HE) 200      & 0.042 & $*$   \\
CLIP vs disSSL $\beta^{-2}$ & Cosine (G$\to$HE) 1000     & 0.015 & $*$   \\

CLIP vs disSSL $\beta^{-1}$ & PCC (HE$\to$G) 200    & 0.028 & $*$   \\
CLIP vs disSSL $\beta^{-1}$ & PCC (HE$\to$G) 500    & 0.032 & $*$   \\

CLIP vs disSSL $\beta^{-1}$ & Cosine (G$\to$HE) 200    & 0.015 & $*$   \\
CLIP vs disSSL $\beta^{-1}$ & Cosine (G$\to$HE) 500    & 0.019 & $*$   \\
CLIP vs disSSL $\beta^{-1}$ & Cosine (G$\to$HE) 1000   & 0.015 & $*$  \\
\bottomrule
\end{tabular}
\end{table*}

\newpage
\subsection*{Training Configurations}

\begin{table}[h!]
\centering
\caption{Training hyperparameters shared across all model configurations.}
\label{tab:hyperparams}
\begin{tabular}{lcc}
\toprule
\textbf{Hyperparameter} & \textbf{VAE-based} & \textbf{Contrastive} \\
\midrule
\multicolumn{3}{l}{\textit{Optimizer}} \\
Learning rate        & $1 \times 10^{-4}$ & $1 \times 10^{-4}$ \\
Weight decay         & $1 \times 10^{-5}$ & --- \\
Optimizer            & Adam (AMSGrad)      & Adam (AMSGrad) \\
\midrule
\multicolumn{3}{l}{\textit{Training schedule}} \\
Batch size           & 64  & 256 \\
Number of epochs     & 50  & 50  \\
\midrule
\multicolumn{3}{l}{\textit{Architecture}} \\
Latent dimension     & 32  & 32  \\
Private latent dim.  & 16  & 16  \\
Encoder model layers &  [512, 256, 64] & [512, 256, 64] \\
\midrule
\multicolumn{3}{l}{\textit{VAE $\beta$-parameter}} \\
$\beta$     & 0.1  & ---  \\
\bottomrule
\end{tabular}%
\end{table}
\newpage

\subsection*{Sample Overview for HEST Colorectal Cancer Cohort}
\begin{table}[h!]
\centering
\caption{Overview of the retained samples used for model training after filtering of low quality samples and enforcing one sample per patient. HEST ID corresponds to the unique identifier of the sample in the HEST database \cite{2jaume2024hest}.}
\label{tab:samples}
\begin{tabular}{lp{8cm}}
\toprule
\textbf{Cohort} & \textbf{HEST IDs} \\
\midrule
Colorectal Cancer    & ZEN37, ZEN40, ZEN42, ZEN43, ZEN44, ZEN47, ZEN49 \\
\bottomrule
\end{tabular}%
\end{table}

\subsection*{Downstream Task Label Generation 
Colorectal Cancer Cohort}
\label{sup:downstream}
 \paragraph{Pathologist Tissue Annotations.} Raw pathologist annotations were obtained from the original publication \cite{2Valdeolivas2024-rr} for each slide. Due to heterogeneity in naming conventions across slides that did not match the categorization criteria, annotations were mapped to a standardized set of tissue region categories. The mapping resolved typographic and formatting variants (normalizing underscores to spaces, converting to lowercase) and merged biologically equivalent labels. Specifically, \textit{connective tissue} classes from slides ZEN37 and ZEN43 were remapped onto corresponding stroma categories based on their immune cell (IC) infiltration grade. \textit{lamina propria}, \textit{squamous epithelium}, and \textit{glandular tissue} were collapsed into \textit{non neoplastic epithelium}. Labels marked as uncertain or appearing in only a single slide were excluded to avoid slide-identity leakage. Each spot was one-hot encoded over the retained categories; unannotated spots were excluded.

 \paragraph{Cell Type Proportion Cluster Labels.} Cell type count estimations from the original publication were downloaded \cite{2Valdeolivas2024-rr}, then normalized to per-spot cell type proportions. To address the compositional nature of these data, proportions were CLR-transformed \cite{2Gloor2017-np} prior to k-means clustering. Clustering was performed once on the pooled, CLR-transformed proportions from all slides jointly, so that a given cluster label corresponds to the same tissue niche across the dataset without requiring post-hoc label harmonization. The optimal k was selected by silhouette score, yielding four tissue niche clusters in both deconvolution estimations (Belgian \& Korean cohort scRNA-seq reference).

 \newpage
 \subsection*{Additional Results on Downstream Tasks }
\begin{figure}[h!]
\includegraphics[width=\textwidth]{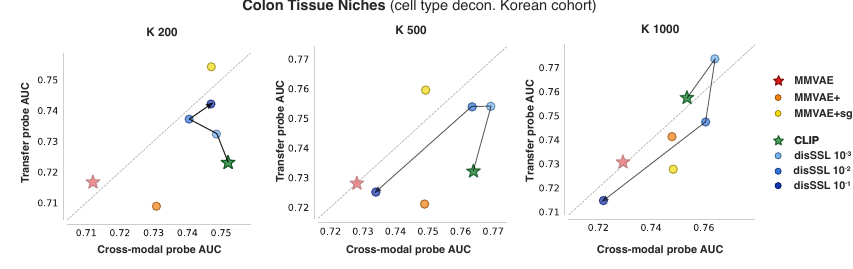}
\caption{\textbf{Downstream task performance for tissue niche classification ($HE \rightarrow G$).}  AUC prediction performance (mean over five repeated runs with different random seeds) from different gene panel sizes $K$ for tissue niche classification derived from scRNA-seq reference Korean cohort. \textit{Stars} indicate the non-disentangled model and \textit{circles} the disentangled versions. The gray arrow connects contrastive models from CLIP ($\beta=0$) to increasing disentanglement strength.} \label{downstream_colon_niches_korean}
\end{figure}

\end{document}